%% file: main.tex
\pdfoutput=1

\documentclass[11pt]{article}

\usepackage[]{acl}
\usepackage{times}
\usepackage{latexsym}

\usepackage[T1]{fontenc}

\usepackage[utf8]{inputenc}

\usepackage{microtype}
\usepackage{amsmath}
\usepackage{xspace}
\usepackage{amsfonts}
\usepackage{enumitem}
\usepackage{cleveref}
\usepackage{graphicx}
\usepackage{multirow}
\usepackage{booktabs}

\usepackage{fixltx2e}

\newcommand{\modelname}{\texttt{TARA}\xspace}
\title{
There's a Time and Place for Reasoning Beyond the Image}

\author{Xingyu Fu$^{1}$, Ben Zhou$^{1}$\thanks{\indent Both authors contributed equally to this work.}, Ishaan Preetam Chandratreya$^{2*}$, Carl Vondrick$^{2}$, Dan Roth$^{1}$ \\
  $^1$University of Pennsylvania \\
  $^2$Columbia University\\
  \texttt{\{xingyuf2, xyzhou, danroth\}@seas.upenn.edu}\\ \texttt{\{ipc2107, cv2428\}@columbia.edu} \\}

\begin{document}
\maketitle
\input{sections/0-abs}
\input{sections/1-intro}

\input{sections/2-related}
\input{sections/3-data-collect}
\input{sections/4-data-analysis}
\input{sections/5-baseline}
\input{sections/7-exp-analysis}

\input{sections/8-conclusion}

\input{sections/8.5-ethic}
\input{sections/9-ack}

\bibliography{anthology,custom}
\bibliographystyle{acl_natbib}
\input{sections/10-appendix}

\end{document}

%% file: sections/0-abs.tex
\begin{abstract}

Images are often more significant than only the pixels to human eyes, as we can infer, associate, and reason with contextual information from other sources to establish a more complete picture. For example, in Figure \ref{fig:new_york_example}, we can find a way to identify the news articles related to the picture through segment-wise understandings of the signs, the buildings, the crowds, and more. This reasoning could provide the time and place the image was taken, which will help us in subsequent tasks, such as automatic storyline construction, correction of image source in intended effect photographs, and upper-stream processing such as image clustering for certain location or time.

In this work, we formulate this problem and introduce \modelname: a dataset with 16k images with their associated news, time, and location, automatically extracted from New York Times\footnote{\url{https://developer.nytimes.com/docs/archive-product/1/overview}} (NYT), and an additional 61k examples as distant supervision from WIT \citep{10.1145/3404835.3463257}. On top of the extractions, we present a crowdsourced subset in which we believe it is possible to find the images' spatio-temporal information for evaluation purpose. We show that there exists a $70\%$ gap between a state-of-the-art joint model and human performance, which is slightly filled by our proposed model that uses segment-wise reasoning, motivating higher-level vision-language joint models that can conduct open-ended reasoning with world knowledge.
The data and code are publicly available at \url{https://github.com/zeyofu/TARA}.

\end{abstract}

%% file: sections/1-intro.tex
\section{Introduction}

\begin{figure}[t]
\centering
\includegraphics[width=1\columnwidth]{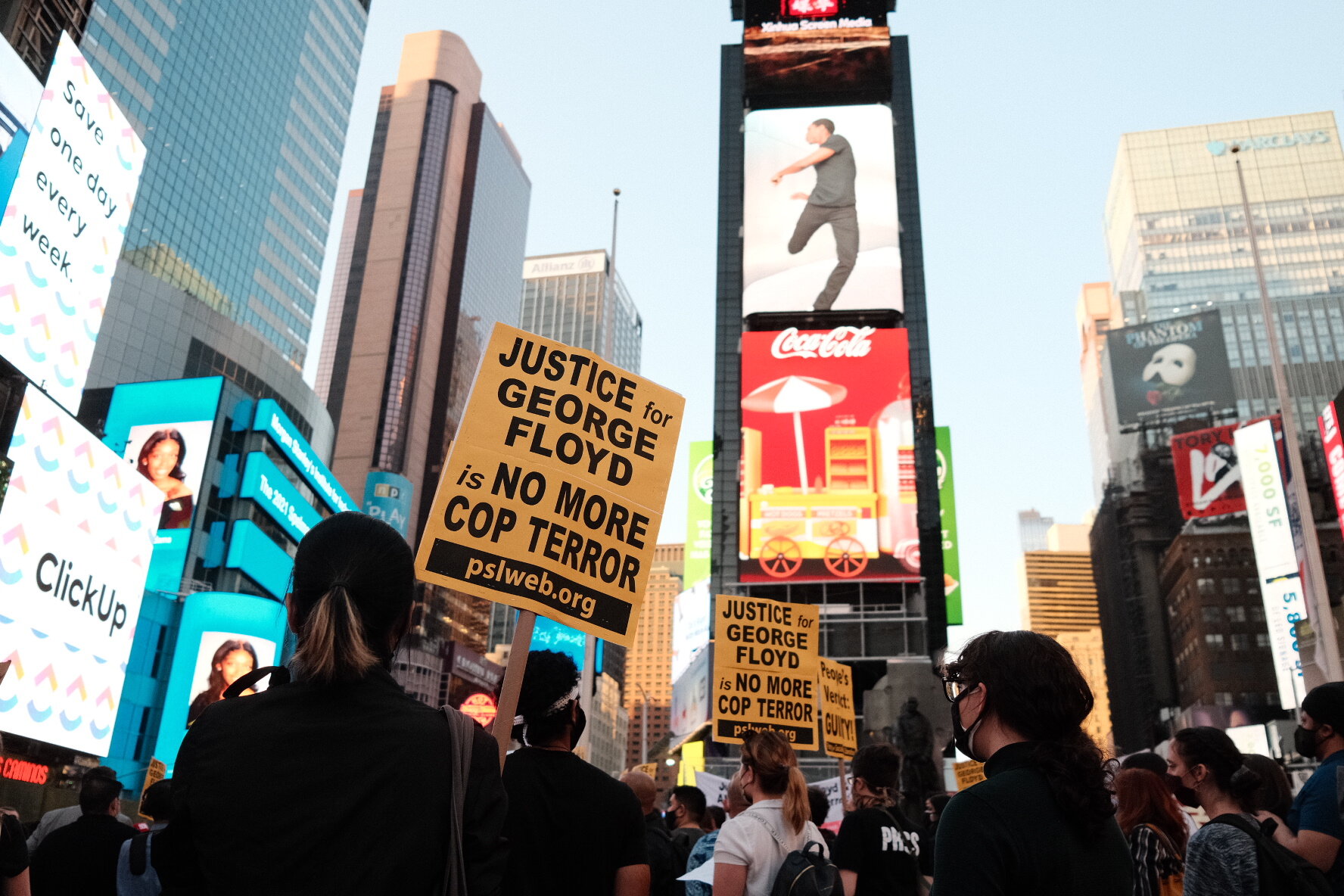}
\vspace{-0.4cm}
\vspace{-0.2cm}
\caption{This is an image from the New York Times. Can you tell the time and location when it was taken?}
\label{fig:new_york_example}
\end{figure}

\begin{figure*}[t]
\begin{center}
    \includegraphics[width=0.5\linewidth]{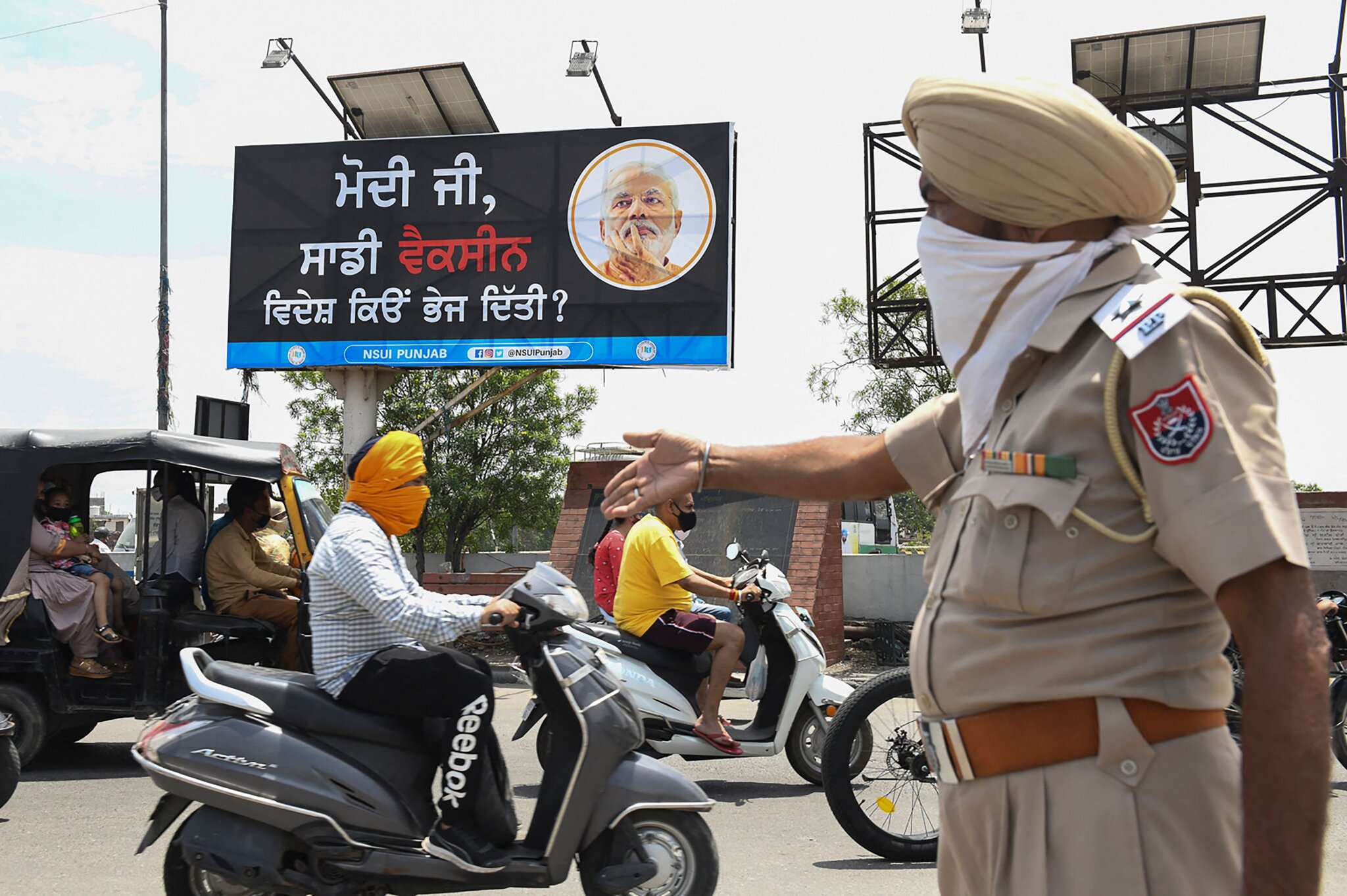}
    \caption{What is the time and location for this image?}
    \label{fig:modi_board}
    \vspace{-1em}
\end{center}
\end{figure*}

\begin{figure*}[t]
\begin{center}
    \hspace*{-0.25cm}
    \includegraphics[width=0.95\linewidth]{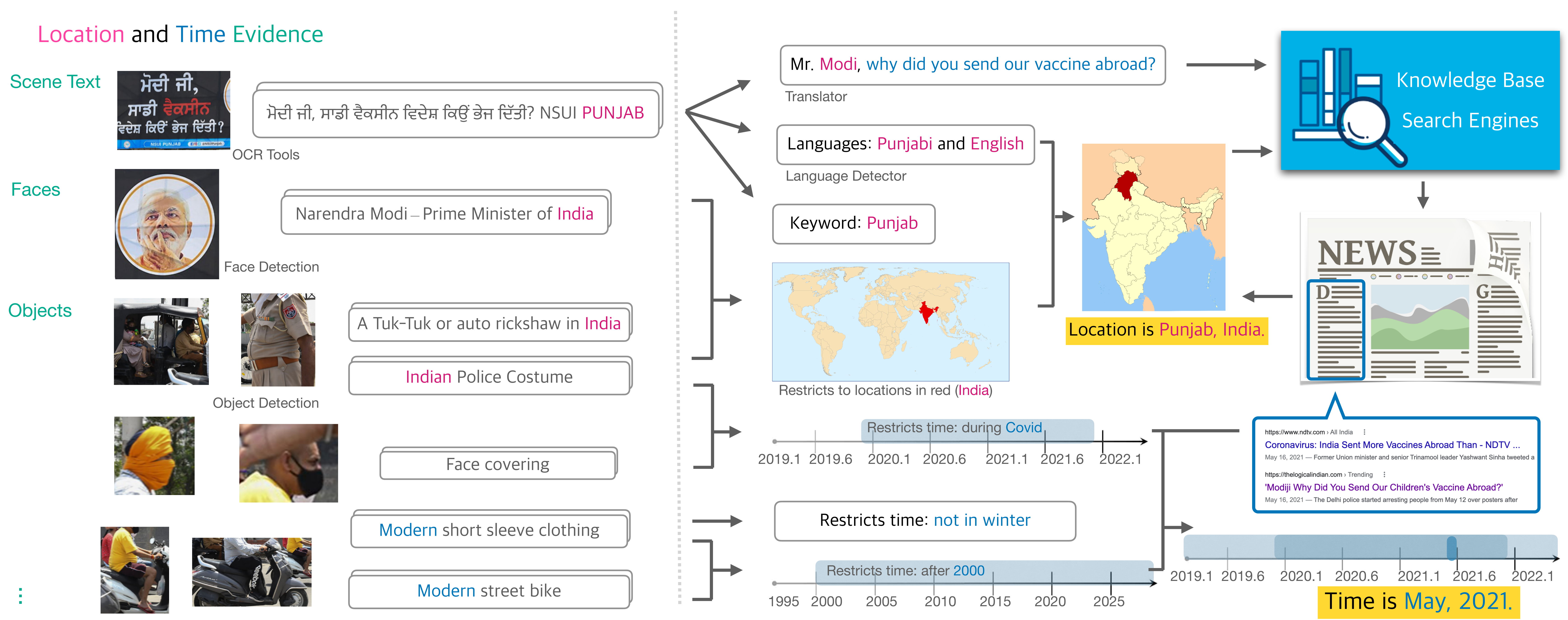}
    \caption{An example of potential joint reasoning on Figure \ref{fig:modi_board} to ground its time and location. Note that people with different backgrounds may need to use different levels of reasoning, resulting in a completely accurate or just partial grounding (e.g., the decade and country), and we only show one such reasoning route.
    We start with grounding multiple scene text, faces, and objects segments from the image, and use the information to conduct a constrained search in a large news-base, until it locates specific textual information related to the image.}
    \label{fig:india_example2}
\end{center}
\end{figure*}

Vision and language are two of most important information sources, and the fact that humans reason jointly with both sources has motivated artificial intelligence research to consider visually-grounded language understanding. Most work in this area has focused on reasoning 
with local evidence \citep{suhr-etal-2019-corpus,hudson2019gqa,Lu_2020_CVPR,liu-etal-2021-visually}, e.g. asking factoid questions such as the colors or shapes of objects and numbers of people, yet very few works \cite{cui2021s} encourage open-ended reasoning where a model needs to look beyond task inputs. 
However, humans can relate visual cues to corresponding contextual information that could be multi-modal, and draw on background knowledge when interpreting and grounding images. 
For example, as Figure \ref{fig:new_york_example} shows, people that are familiar with the news can infer that the location is Times Square through the iconic screen panels, and further estimate the period of time by looking at the crowds and the signs. And, this can be done without explicitly including related news pieces as input. 
In fact, even though some people would not have the prior knowledge to identify the relevant events, it is likely that they would have good estimate of the location and time by interpreting textual evidence in the image, the language in the signs, entity names, building styles, and other details in the input image.

In this work, we identify and formulate this problem, spatio-temporal grounding of images, a task aiming at identifying the {\em time} and {\em location} in which the given image was taken.  
Specifically, we develop a novel dataset \modelname, (\underline{T}ime and pl\underline{A}ce for \underline{R}easoning beyond the im\underline{A}ge), a challenging dataset that tasks models with grounding images to real-world spatial and temporal information. In our collection, we make sure that for models to accurately find images' creation time and location, they would need to successfully ground the visual clues in
texts such as news, stories and encyclopedias. As a result, this task motivates models to consider the association between visual information, language, and background knowledge, more closely and in a more open-ended setting. Figure \ref{fig:modi_board} shows an example from \modelname, and Figure \ref{fig:india_example2} shows a possible way for a model to ground the image to its spatio-temporal information. The system starts with grounding multiple segments from the image, and uses the information to conduct a constrained search in a large news-base, until it locates specific textual information related to the image. This demonstrates the complexity and significance of this task.

\modelname{} is collected via a rigorous process that involves rule-based distant supervision extraction from news-images data which results in 16k image examples. While the training data has high label correctness (around 95\%), we further run a crowdsourced validation on 3k examples to form the evaluation dataset. During the validation, annotators are asked to verify that there exists a potential path for humans to derive the correct answer, which encourages proper reasoning in future works. To better support the study of domain transfer and supervision for the reasoning process, we collect an additional 61k examples from the Wikipedia domain. We apply the state-of-the-art joint model CLIP \citep{radford2021learning} and show that it only achieves accuracy of 11.11\% and 0.46\% for time and location, respectively, on our dataset. 
Additionally, we present a new CLIP-based baseline model that reasons on object and facial segments and achieves 16.46\% and 1.07\% accuracy for time and location, respectively. 
We show that there exists a large gap (around 70\% in accuracy) between state-of-the-art models and human performance, suggesting that the \modelname data will provide a benchmark to motivate reasoning based approaches and support significant future work.

%% file: sections/2-related.tex
\section{Related Work and Datasets}

\textbf{Vision and Language Learning}
Language understanding in the context of images has been widely studied in various datasets covering a wide range of tasks including visual question answering, image retrieval, image and video captioning, etc.
Earlier datasets mostly focus on simple local object properties identification \citep{antol2015vqa,chen2016semi}.
Later on, datasets start to focus on compositional visual reasoning. For example, \citet{suhr-etal-2017-corpus} and \citet{johnson2017clevr} use synthetic images or synthetic language to study spatial relations.
Recently, datasets using real images and real languages such as \citep{hudson2019gqa,liu-etal-2021-visually} were proposed for reasoning about natural language descriptions of photos.
However, all of the datasets focus on local grounding on segments inside the image, but not globally ground beyond the image with open-ended reasoning.

While there are various tasks and datasets, the underlying associations between language and visual concepts are often common across different tasks \citep{Lu_2020_CVPR}.
Therefore, we use CLIP \citep{radford2021learning} to study the \modelname dataset in this paper.
CLIP is a recently released state-of-the-art image representation model which has shown impressive performance on various tasks through pre-training on 400 million image and captions pairs collected from the internet.

\noindent\textbf{Spatio-temporal IE from Texts}
There has been extensive work on identifying temporal expressions and their associations with events in texts. \citet{UzZaman2013SemEval2013T1, ning-etal-2018-cogcomptime} focus on temporal information extraction within the local contexts, and \citet{zhou-etal-2020-temporal, zhou-etal-2021-temporal} further extends the scope to consider contextual information from external texts.
The NLP community has also investigated spacial information extraction, with geocoding \cite{Gritta2018WhatsMI, Kulkarni2020SpatialLR}, which maps mentions to geological coordinates, being closest to our scope. 

%% file: sections/3-data-collect.tex
\section{Dataset Collection}

\begin{figure*}[t]
\begin{center}
    \includegraphics[width=1\linewidth]{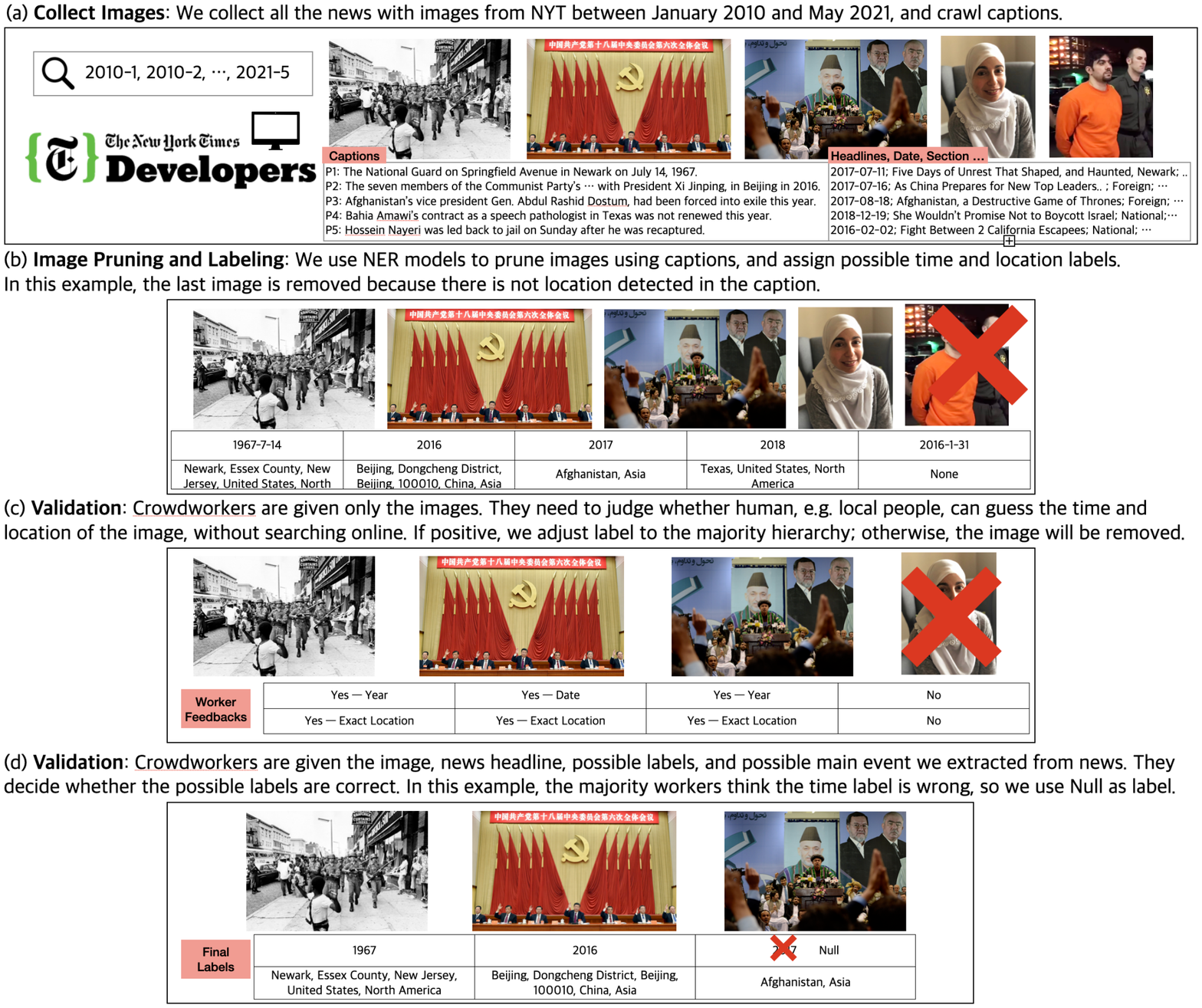}
    \caption{Data collection process. Steps (a)–(b) are described in Section \ref{sec:data-image}; and steps (c)-(d) in Section \ref{sec:data-validation}.}
    \label{fig:data}
    \vspace{-1em}
\end{center}
\end{figure*}

Each example in \modelname includes a news image, along with its time, location, caption, and corresponding news background such as headline, abstract, and news type. These are included for training or analysis purposes, but the task is to guess the correct time and location as accurately as possible given only the image.
In developing the dataset, our goal is to collect a large corpus of semantically rich images that human with world knowledge can correctly identify the time and location, using evidence from the image, background knowledge, and appealing to external knowledge (which we call ``reasoning" here).
We design the process of collecting and identifying the images so that it facilitates this type of reasoning, and then use crowd sourcing to label a random 20\% of high-quality images for development and testing. 
Figure \ref{fig:data} illustrates our data collection procedure. 

\subsection{Image collection}
\label{sec:data-image}
We first collect all the news between January 2010 and May 2021 using the NYT API~\footnote{\url{https://developer.nytimes.com/docs/archive-product/1/overview}}.
We did not collect news that are earlier than 2010 because earlier news articles contain much fewer images. 
Each news article comes with a list of attributions\footnote{For each news, the API provides attributes as listed here: \url{https://developer.nytimes.com/docs/archive-product/1/types/Article}} such as headline, abstract, news type, and possibly a main image. 
We first filter the news articles that has a valid image, and then scrape image caption for each image. 
Since the NYT covers news in several multimedia formats, the images follow a range of formatting practices, such as representative news images, image collages, images sampled from slideshows and descriptive natural thumbnails for videos.
We setup a NYT specific pipeline to scrape image captions. We define a separate scraping procedure to get image specific text information for the different media types mentioned above and remove instances where multiple and/or ambiguous captions are returned. 

\textbf{Image Pruning and Labeling }
Next, we describe how we automatically collect time and location of an image from corresponding news articles and captions.
First, we filter out the images with unwanted news types such as reviews, series, and obituaries, and unwanted news topics such as food, fashion, and movies, because images from these articles may not be informative enough. 
Then, we filter out the images whose caption does not contain location and time. For those that contain temporal and spacial cues, we assign each image a possible time label and location label.
Specifically, we use the Spacy NER model\footnote{\url{https://spacy.io/models/en}} to find if the caption has both exactly one ``DATE'' entity for time and one ``GPE'' or ``LOC'' typed entity for location. 
Note that each news comes with a publication date and possible locations in attributes. 
We would either directly use our NER-extracted time entity as the possible time label if it's a valid time, or adjust the publication date using the time entity. For example, if the time entity is ``1936'' and publication date is ``2021-05-01'', then we will use ``1936'' as the possible time label because it should be an old image occurring in a recent news; in the latter case, if the time entity is ``last month'' and publication date is ``2015-07-18'', then we will use ``2015-06'' as the possible time label. 
We also compare our NER-extracted location entity with the news attribute locations. If the only difference is granularity, e.g. one is New York, United States and the other is United States, then we will use the fine-grained one ``New York, United States'' as possible location label. Otherwise, we will filter our this image.

Finally, we add missing hierarchies for each possible label. For time labels, we add the decade and the century. 
For location labels, we use Geopy\footnote{\url{https://geopy.readthedocs.io/en/stable/}} to identify the location and add missing hierarchies such as country and continent.

\subsection{Validation}
\label{sec:data-validation}
We randomly select an equal number of images from each month, such that a total of about 20\% images are assigned to devlopment and test. 
On these images, we use two crowdsourcing tasks to (1) prune unanswerable images, and (2) verify correctness of the labels.

In the first task, we display a single image, and ask a worker to answer, without searching online, if any person can guess the time and location of the image. We offer different hierarchies in the choices -- date, year, decade, and century for time and exact location, city, country, and continent for location -- so that workers can choose one of these. 
If the majority of workers agree that human cannot reason time or location based on the image itself, we will mark the corresponding label as null.
Otherwise, if the majority of them agree on a certain hierarchy, we adjust the possible label to that specific hierarchy.
Check step(c) in Figure \ref{fig:data} for criteria and positive and negative examples. 

The second task further verifies the correctness of current time and location labels. Specifically, we provide the same image, but including its caption, news headline, abstract, and extracted time and location labels. 
We ask the workers to verify if the background event is the same as in image, and if the labels are correct after reading the additional information.
We use the Semantic Role Labeling (SRL) model\footnote{\url{https://demo.allennlp.org/semantic-role-labeling}} from AllenNLP to detect the main verb in the image caption by selecting the verb with most arguments, and mark it as the possible main event to provide to the workers. Detailed examples can be found in step(d) in Figure \ref{fig:data}.

\begin{figure*}[t]
\begin{center}
    \includegraphics[width=1\linewidth]{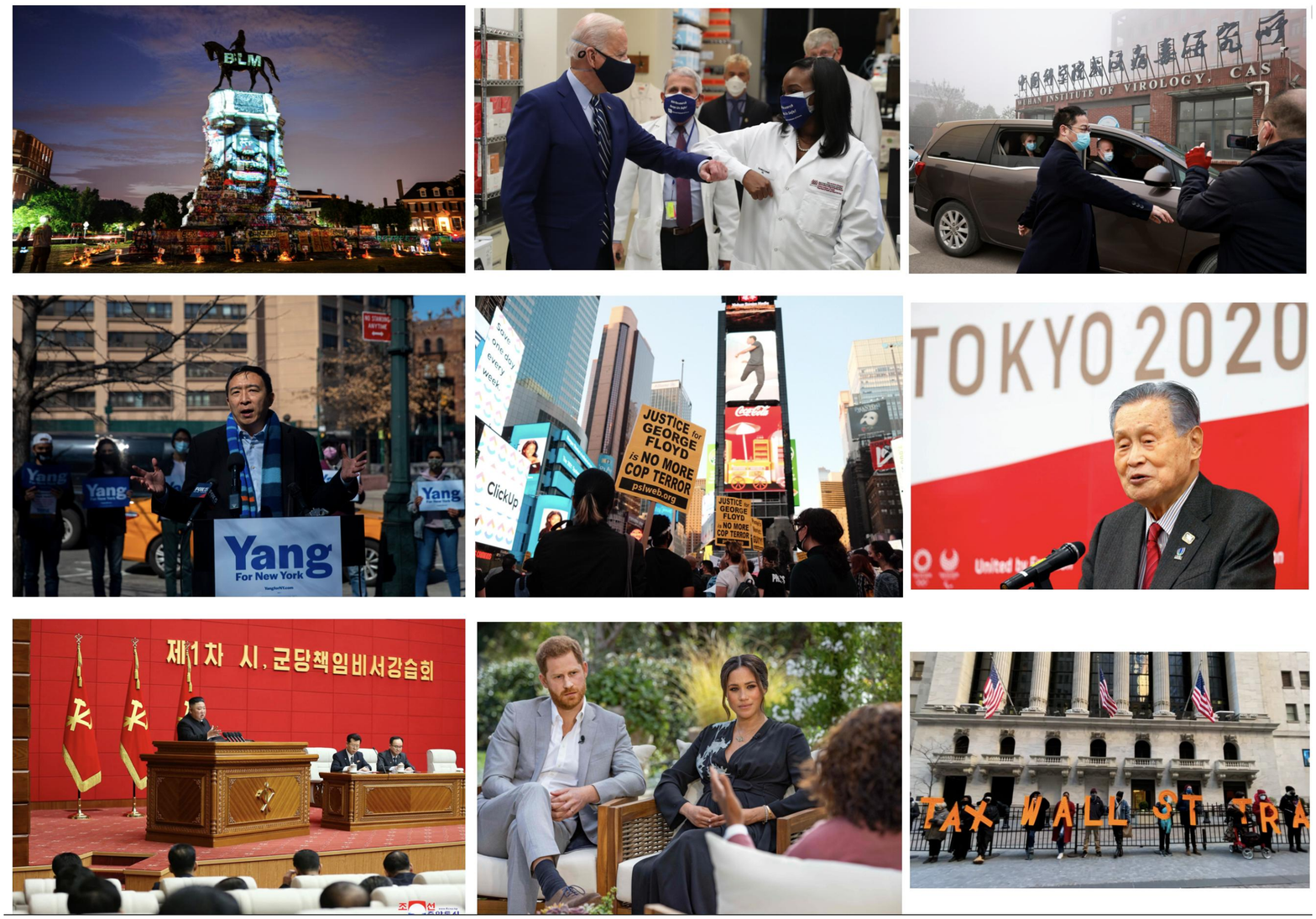}
    \caption{Some example images in our test set of interest as described in Section \ref{sec:data-interest}. These very recent images require open-ended reasoning with world knowledge and are specifically chosen such that our human baseline annotators probably have enough knowledge about the key evidence. For example, in the first image, people need to know what ``BLM'' is so that they can start to search statues in United States. Also in the second image, people need to know it is the President Biden for further reasoning.
    }
    \label{fig:data-interest}
    \vspace{-1em}
\end{center}
\end{figure*}

\subsection{Test Set of Interest} \label{sec:data-interest}
We further select a small set of 30 interesting images as shown in Figure \ref{fig:data-interest}, that are related to most famous news happening after January 2021, the CLIP model date.\footnote{\url{https://github.com/openai/CLIP/blob/main/model-card.md}}
This adversarial test set is specifically chosen to cover unseen images by baseline models to better test their generalization instead of memorization.

Additionally, regarding to human baseline, annotators need to have enough knowledge to extract and interpret the key evidence segments, in order to reason about the answer. 
For instance, a person with an American cultural background and speaks English but not Hindi may find Figure \ref{fig:new_york_example} is easier to infer the precise time and location than Figure \ref{fig:modi_board}, compared to a person with Indian cultural background and speaks Hindi but not English, and vise versa.
This test set of interest is chosen to cover most well-known news for the purpose that human baseline annotators are more likely to have enough knowledge about the key evidence so that the comparison with neural models can be more fair.

\subsection{Additional Weak Supervision} 
We apply the same image pruning and labeling procedures on the WIT dataset~\citep{10.1145/3404835.3463257}, which contains 11.5M Wikipedia images and the surrounding paragraphs and captions.
Since this dataset is much unorganized, we only select images in English Wikipedia articles, and apply two additional NER models~\citep{lample-etal-2016-neural,peters-etal-2017-semi} from AllenNLP\footnote{\url{https://demo.allennlp.org/named-entity-recognition/fine-grained-ner}} to select locations.
We further use zero-shot CLIP model to prune unwanted image types. Specifically, we provide each image with text sentences in the format of ``a photo of [\textit{type}]'', with \textit{type} being \textit{photograph}, \textit{map}, \textit{paint}, and \textit{paper}, and retrieve the sentence with highest similarity score.
We only keep images of type \textit{photograph}, and use these as additional weak supervision. 
The benefit of adding this additional weak supervision is that it has a wider range of time and location labels than the NYT images, especially because that all the NYT images are taken from news between 2010 and 2021.

%% file: sections/4-data-analysis.tex
\section{Dataset Analysis}

\begin{figure*}[t]
\begin{center}
    \hspace*{-1.5cm}
    \includegraphics[width=1\linewidth]{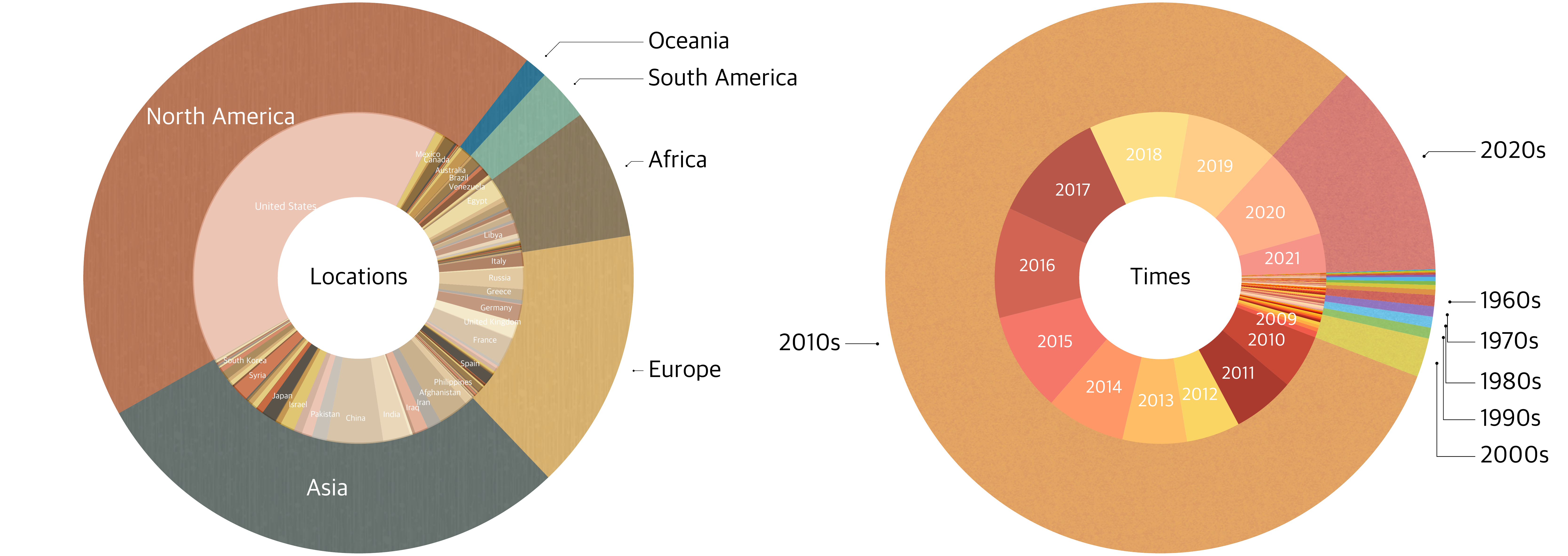}
    \caption{Label distribution in \modelname. All of the training, development, and testing data are considered.}
    \label{fig:dist}
    \vspace{-1em}
\end{center}
\end{figure*}

\subsection{Dataset Statistics}

\begin{table}[t]
\small
\centering
\setlength{\tabcolsep}{4pt}
\begin{tabular}{c|cccc}
\toprule
Dataset & Train & Dev & Test & All\\ 
\midrule
\modelname before validation & 12,306 & 1,644 & 1,644 & 15,652 \\
\modelname & 12,306 & 1,552 & 1,571 & 15,429 \\
\midrule
WIT &  &  &  & 61,325 \\
\bottomrule
\end{tabular}
\vspace{-0.5em}\caption{Dataset statistics for \modelname and additional WIT supervision.}
\label{tab:stat}
\vspace{-1em}
\end{table}

Dataset statistics can be found in Table \ref{tab:stat}.
\modelname contains about 16K images from New York Times. 
After crowd-sourcing validation on development and testing, about 94\% of the images that either has a valid location label or time label are kept, indicating that our training set can serve as a good weak supervision. 
In addition, \modelname provides a 61K weak supervision dataset built upon WIT. 

\subsection{Time and Location Distribution}
\Cref{fig:dist} shows the time and location distribution in \modelname. We can see that most images are taken in North America, Asia, and Europe, between 2010 and 2021.
This can be the effect of using NYT as image source.

%% file: sections/5-baseline.tex
\section{Baselines}
We assess the quality of our dataset through human annotation, and evaluate on existing visual reasoning approaches. 

\subsection{Human Performance}
As introduced in Section \ref{sec:data-interest}, an expert annotator works on our test set of interest to gain a better understanding of the human performance on \modelname. The expert is not allowed to directly search the image online, but can search for anything else such as the keywords she/he infers from the image.
The expert is presented with all the labels in the test set just as neural models. 

\subsection{Evaluation Systems}
We use the state-of-the-art systems
in machine reading comprehension for this task:
\textbf{CLIP}~\citep{radford2021learning}.
CLIP is the state-of-the-art image representation model and has shown impressive progress on visually grounded language understanding tasks. 
Specifically, we use the ``ViT-B/32'' model\footnote{\url{https://github.com/openai/CLIP}} for zero-shot classification and analysis.
During prediction, the model is given a single image and needs to classify the correct label. We use a similar prompt template ``A photo taken in \{label\}.'' following the original paper, to encode all the labels. We compare the similarity between the image and each label prompt, and the highest one is the predicted label.

We also add several variants of CLIP. The first is \textbf{CLIP+}, which is the zero-shot CLIP model fine-tuned on NYT training data.
Note that CLIP uses contrastive loss to train on image and text pairs.
We concatenate the time and location labels into a natural language sentence to serve as the text part for an image.

\textbf{CLIP+Seg} is another variant where we first extract object and face segments, and then finetune the CLIP model on the whole images along with the segments, both with time and location labels concatenated together as the final goal.
As for object detection, we use the YOLOv5\footnote{\url{https://github.com/ultralytics/yolov5}} method, specifically with model ``yolov5s''.
The intuition is that for objects such as iPhone, the model benefits from training it to times later than 2010.
We add a limit to the segments so that we only consider important objects that have size larger than 50. We further restrict the number of people segments to be no more than 3, since many of the images have crowds and adding more people do not bring in much additional information.
As for face segments, we use the InsightFace~\citep{guo2022sample} facial detection model\footnote{\url{https://github.com/deepinsight/insightface}}.
The intuition is that for famous people such as President Biden, we will benefit from training the segments to location ``United States''.
During implementation, we also add a limit to the segments so that we only consider face that have size larger than 50, which are more likely to be most important faces. 

\textbf{CLIP+WIT} is the variant of CLIP where we finetune on the training images along with the 61K weak supervision Images extracted from WIT. We concatenate the possible time and location labels as the paired text.

%% file: sections/7-exp-analysis.tex
\section{Experiments and Results}

\subsection{Evaluation metrics}
Two metrics are adopted in this work: Accuracy and
\textbf{Example-F1} (also known as micro-Dice coefficient) following previous studies \citep{shen-etal-2021-taxoclass}. 
Accuracy is calculated without considering hierarchies -- the predicted label needs to exactly match the gold label.
In contrast, \textbf{Example-F1} calculates the average F1 scores considering each hierarchy as follows:

\begin{equation}
\label{eq1}
\text { Example-F1 }=\frac{1}{N} \sum_{i=1}^{N} \frac{2\left|\mathbb{L}_{i}^{\text{true }} \cap \mathbb{L}_{i}^{\text{pred }}\right|}{\left|\mathbb{L}_{i}^{\text{true }}\right|+\left|\mathbb{L}_{i}^{\text{pred }}\right|}
\end{equation}
where $\mathbb{L}_{i}^{\text {true }}$($\mathbb{L}_{i}^{\text {pred }}$) is the true (model predicted) hierarchical label set of image $i$.
For example, if the true labels for an image are ``1967-7-14'' and ``Newark, New Jersey, United States, North America'' respectively, then its true hierarchical label sets are [``Newark, New Jersey, United States, North America'', ``United States, North America'', ``North America''] and [``1967-7-14'', ``1967-7'', ``1967'', ``1960s'', ``20th century''].

\subsection{Experimental results}

\input{files/results}
In Table \ref{tab:result}, we report the experimental results using the CLIP based baselines on the \modelname. 
We can see that all of the model performance still have a large gap with human performance.
Also, the object and facial segments boosts the model to be the highest on location prediction, proving that segment level reasoning is needed in this task.
In contrast, adding the WIT weak supervision does not show consistent improvement or reduction on the performance.
It can be due to that WIT images are not similar to news images, and that WIT images are mostly taken in older times than 2010, thus not providing enough supervision for our test set.
There is also an obvious gap between the location prediction and time prediction, showing that temporal reasoning in vision language learning is much under explored and needs further research.
Note that the Example-F1 value is consistently higher than accuracy because if the model predicts the highest two hierarchies correctly (e.g. century and decade), then it gets an Example-F1 around 40\%.

\subsection{Analysis}
\label{sec: analysis}
We perform qualitative and quantitative analysis of the baseline results to better understand the strengths and weaknesses of CLIP based models, and hypothesize avenues for future work.
Specifically, we look into: model performance on test set of interest; effects on performance by using news abstract.

\input{files/results_interest}
\textbf{Test Set of Interest}
Since we conduct human evaluation only on the test set of interest, we examine how models perform on this set and show the results in Table \ref{tab:result_interest}.
Note that we use the same setting for the models and human experts -- both are given the entire test set labels.
From the results, we observe a large gap between between the model performance and human performance, indicating that existing sota model still lacks a certain level of reasoning capability required to solve a such hard task as defined in the \modelname dataset.
Comparing the results in Table \ref{tab:result_interest} to those in Table \ref{tab:result}, we can see that there is little performance difference for each model, indicating that our human performance on the test set of interest can serve as a good reference to human performance on the whole test set, under the assumption that the annotators have enough knowledge about the key evidence segments.

\input{files/results-news}
\textbf{News Abstracts} We also experiment with news abstracts being the classification goal instead of time and location labels given an image, under the assumption that models are given corresponding news abstract for each label.
The intuition is that the news abstract might provide more descriptions that can map to several local segments, and thus providing additional information.
Comparing the results shown in Table \ref{tab:result_news} to Table \ref{tab:result}, we can see that providing news abstracts improves the performance a lot, despite that there is still a large gap with human performance.


%% file: files/results.tex
\begin{table}[t]
\small
\centering
\setlength{\tabcolsep}{4pt}
\begin{tabular}{c|cc}
\toprule
Model & Accuracy & Example-F1
\\ \midrule
CLIP & 11.11 & 44.96 \\
CLIP+ & 15.72 & 49.74 \\
CLIP+WIT & 11.11 & 45.20 \\
CLIP+Seg & \textbf{16.46} & \textbf{50.52} \\
\midrule
Human & 86.21 & 92.41 \\
\bottomrule
\end{tabular}
\vspace{0.5em}

\begin{tabular}{c|cc}
\toprule
Model & Accuracy & Example-F1
\\ \midrule
CLIP & 0.46 & 39.90 \\
CLIP+ & 1.00 & \textbf{43.09} \\
CLIP+WIT & \textbf{1.07} & 41.73 \\
CLIP+Seg & 0.92 & 42.82 \\
\midrule
Human & 75.86 & 91.63 \\
\bottomrule
\end{tabular}
\vspace{-0.5em}\caption{Summary of the performance(\%) for different baselines on the image location prediction (above) and time prediction (bottom). 
Definition of Example-F1 is in Equation \ref{eq1}. Note that human performance here is evaluated on the test set of interest instead of on the whole test set, please see Section \ref{sec: analysis} for more details.
}
\label{tab:result}
\vspace{-1em}
\end{table}

%% file: files/results_interest.tex
\begin{table}[t]
\small
\centering
\setlength{\tabcolsep}{4pt}
\begin{tabular}{c|cc}
\toprule
Model & Accuracy & Example-F1
\\ \midrule
CLIP & 13.33 & 56.44 \\
CLIP+ & 13.33 & 58.67 \\
CLIP+WIT & 10.00 & 55.11 \\
CLIP+Seg & \textbf{23.33} & \textbf{63.11} \\
\midrule
Human & 86.21 & 92.41 \\
\bottomrule
\end{tabular}
\vspace{0.5em}

\begin{tabular}{c|cc}
\toprule
Model & Accuracy & Example-F1
\\ \midrule
CLIP & 0.00 & 24.65 \\
CLIP+ & 0.00 & 26.49 \\
CLIP+WIT &0.00 & \textbf{29.83} \\
CLIP+Seg & \textbf{3.33} & 24.43 \\
\midrule
Human & 75.86 & 91.63 \\
\bottomrule
\end{tabular}
\vspace{-0.5em}\caption{Performance(\%) of different baselines evaluated on the test set of interest for image location prediction (above) and time prediction (bottom). }
\label{tab:result_interest}
\end{table}

%% file: files/results-news.tex
\begin{table}[t]
\small
\centering
\setlength{\tabcolsep}{4pt}
\begin{tabular}{c|cc}
\toprule
Model & Accuracy & Example-F1
\\ \midrule
CLIP & \textbf{28.18} & 61.63 \\
CLIP+ & 26.49 & \textbf{62.68} \\
CLIP+WIT & 11.11 & 50.00 \\
CLIP+Seg & 26.96 & 62.41 \\
\bottomrule
\end{tabular}
\vspace{-0.5em}\caption{Performance(\%) for different baselines predicted towards news abstracts.}
\label{tab:result_news}
\vspace{-1em}
\end{table}

%% file: sections/8-conclusion.tex
\section{Conclusion}
In this work, we introduce \modelname, a new dataset and task for spatio-temporal grounding of images that requires open-ended joint reasoning with world knowledge.
\modelname provides a dataset of 16K high-quality images from NYT and Wikipedia-based supervision for additional 61K images. 
Compared to previous visual-language understanding datasets, \modelname requires more complicated reasoning ability and existing state-of-the-art models such as CLIP are far from human levels, suggesting that our task remains a significant challenge with large room for improvement.
We hope that \modelname will inspire future work on reasoning beyond image's local segments in vision-language understanding.

%% file: sections/8.5-ethic.tex
\section{Ethical Considerations}

We collected data for \modelname by downloading raw data from the official NYT API at \url{https://developer.nytimes.com}. 
According to the Terms of Use at \url{https://developer.nytimes.com/terms} and NYTimes.com Terms of Service located at \url{https://help.nytimes.com/hc/en-us/articles/115014893428-Terms-of-service}, NYT granted us a license to access the NYT APIs and scrape their data.
We ensure that our dataset has been collected in a manner which is consistent with the terms of use of NYTimes.

We only release our dataset \modelname for academic purpose.
In order to retrieve the same raw data we scraped from the NYT API, multiple requests for months between January 1, 2010 and May 31, 2020 need to be made following the instructions at \url{https://developer.nytimes.com/docs/archive-product/1/overview}.

As introduced in Section \ref{sec:data-validation}, we annotated the data using crowd-workers through Amazon Mechanical Turk.
They are voluntary participants who were aware of any risks of harm associated with their participation.
We require the workers to be located in either Australia, Canada, Great Britain or the United States such that they are English speakers. We also require the workers to have HIT Approval Rate (\%) for all Requesters' HITs greater than or equal to 98\%. 
All crowd-workers were compensated by a fair wage determined by estimating the average completing time of each annotation task. Each worker earn \$2.4 per 10 queries and each query should take less than a minute to annotate. 
Example screenshots of the NYT data and our annotation interface can be found in Appendix \ref{sec:appendix-mturk}.

%% file: sections/9-ack.tex
\section*{Acknowledgments}
This research is based upon work supported in part by the office of the Director of National Intelligence (ODNI), Intelligence Advanced Research Projects Activity (IARPA), via Contract No. 2019-19051600006 under the BETTER Program, by Contracts FA8750-19-2-1004 and FA8750-19-2-0201 with the US Defense Advanced Research Projects Agency (DARPA), and by a grant from the Allen Institute for Artificial Intelligence (allenai.org).
The views and conclusions contained herein are those of the authors and should not be interpreted as necessarily representing the official policies, either expressed or implied, of ODNI, IARPA, the Department of Defense, or the U.S. Government. 

%% file: sections/10-appendix.tex
\appendix

\section{Example Screenshots from the NYT Website and MTurk Annotation Interface}
\label{sec:appendix-mturk}
\begin{figure}[t]
\centering
\includegraphics[width=1\columnwidth]{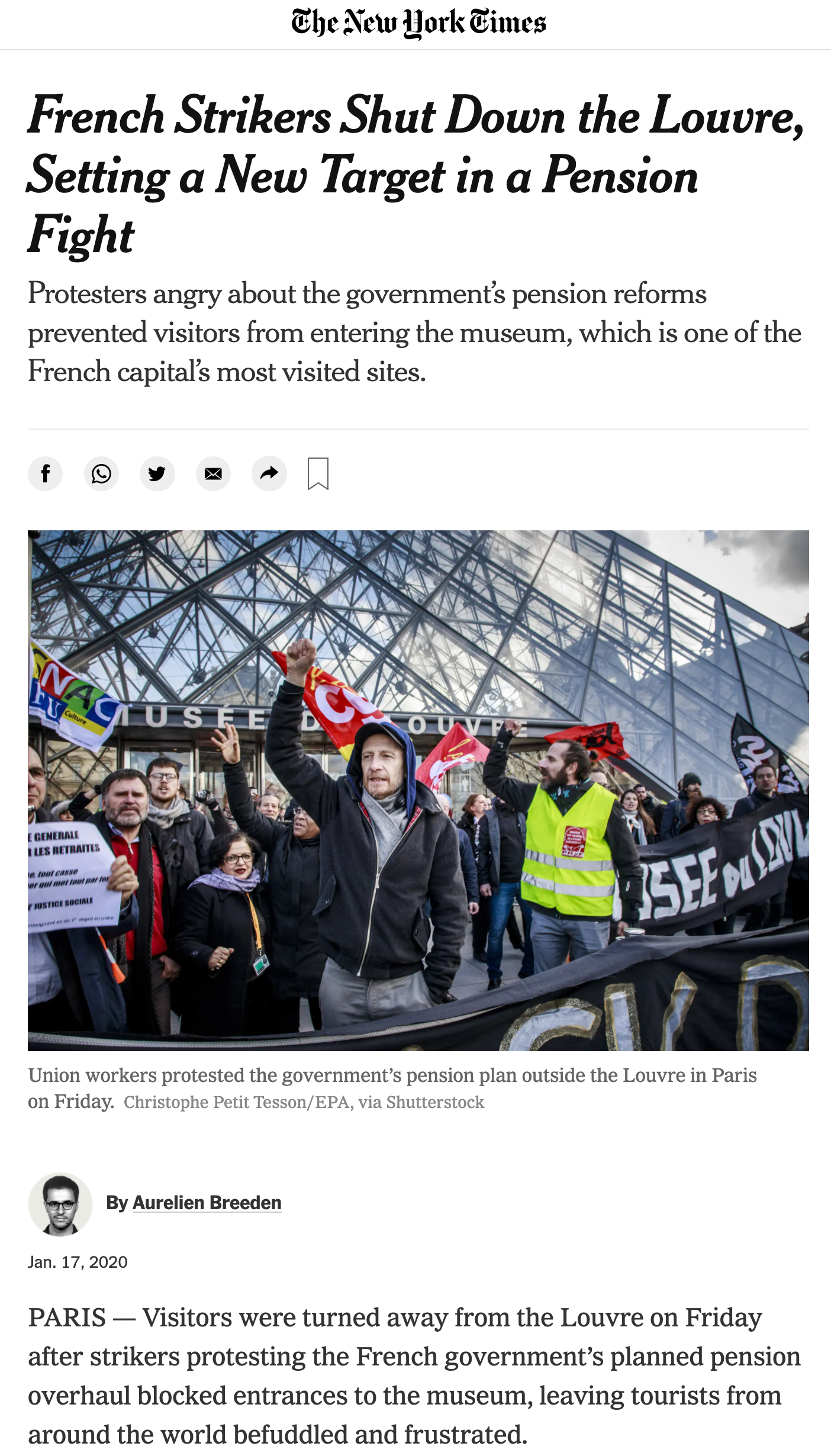}
\vspace{-0.4cm}
\vspace{-0.2cm}
\caption{This is an example news from the New York Times(NYT) website.}
\label{fig:nyt_example}
\end{figure}

\begin{figure*}[t]
\begin{center}
    \includegraphics[width=1\linewidth]{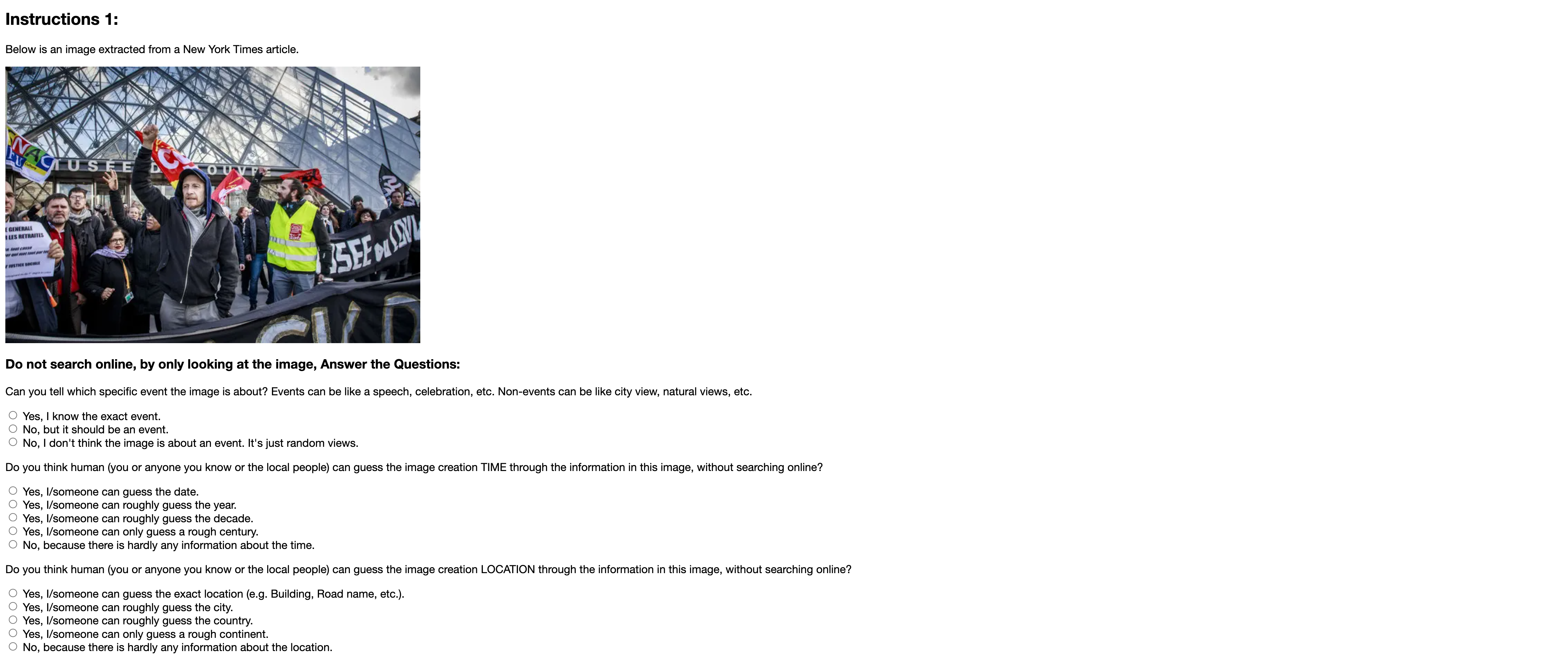}
    \vspace{-1em}
    
    \includegraphics[width=1\linewidth]{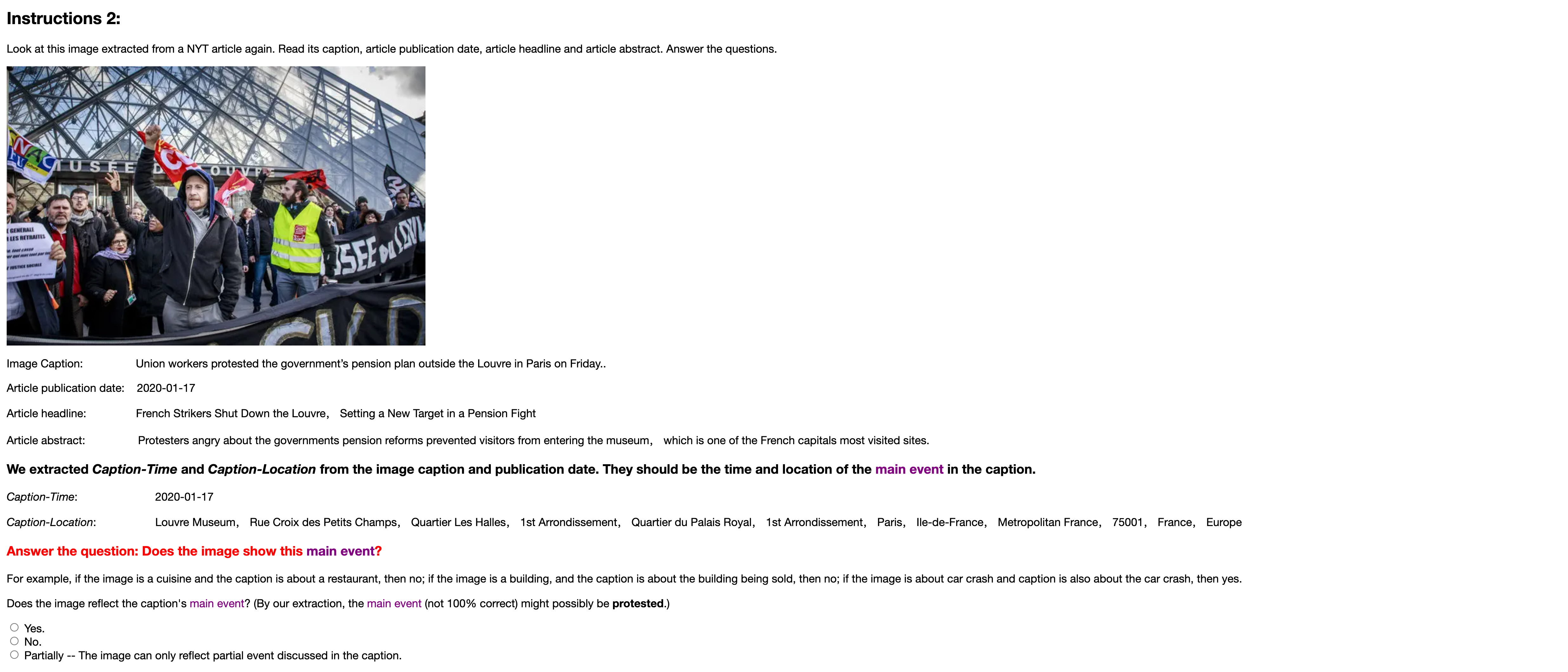}
    \vspace{-1em}
    
    \includegraphics[width=1\linewidth]{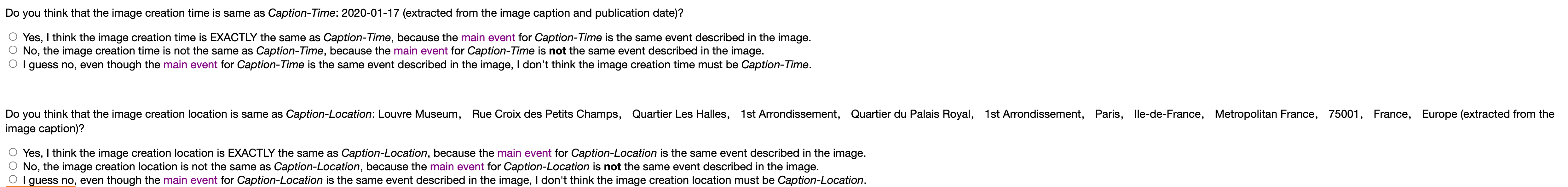}
    \caption{A screenshot of the MTurk annotation instructions for data validation as introduced in Section \ref{sec:data-validation}.}
    \label{fig:mturk}
    \vspace{-1em}
\end{center}
\end{figure*}

In this section, we first show an example news screenshot taken from the NYT website located at \url{https://www.nytimes.com/}, where we use the provided API to download the data, as in Figure \ref{fig:nyt_example}. We then show example screenshot of our data annotation process as described in Section \ref{sec:data-validation}. 
For the data annotation tasks, we present the Turkers with step-by-step instructions of the tasks that we require them to do, along with carefully selected examples. More details can be found in Figure \ref{fig:mturk}.

%% file: main.bbl
\begin{thebibliography}{21}
\expandafter\ifx\csname natexlab\endcsname\relax\def\natexlab#1{#1}\fi

\bibitem[{Antol et~al.(2015)Antol, Agrawal, Lu, Mitchell, Batra, Zitnick, and
  Parikh}]{antol2015vqa}
Stanislaw Antol, Aishwarya Agrawal, Jiasen Lu, Margaret Mitchell, Dhruv Batra,
  C~Lawrence Zitnick, and Devi Parikh. 2015.
\newblock {VQA}: Visual question answering.
\newblock In \emph{Proceedings of the IEEE international conference on computer
  vision}, pages 2425--2433.

\bibitem[{Chen et~al.(2016)Chen, Lucchi, and Hofmann}]{chen2016semi}
Wenhu Chen, Aurelien Lucchi, and Thomas Hofmann. 2016.
\newblock A semi-supervised framework for image captioning.
\newblock \emph{arXiv preprint arXiv:1611.05321}.

\bibitem[{Cui et~al.(2021)Cui, Khandelwal, Artzi, Snavely, and
  Averbuch-Elor}]{cui2021s}
Yuqing Cui, Apoorv Khandelwal, Yoav Artzi, Noah Snavely, and Hadar
  Averbuch-Elor. 2021.
\newblock Who's {W}aldo? linking people across text and images.
\newblock In \emph{Proceedings of the IEEE/CVF International Conference on
  Computer Vision}, pages 1374--1384.

\bibitem[{Gritta et~al.(2018)Gritta, Pilehvar, Limsopatham, and
  Collier}]{Gritta2018WhatsMI}
Milan Gritta, Mohammad~Taher Pilehvar, Nut Limsopatham, and Nigel Collier.
  2018.
\newblock What’s missing in geographical parsing?
\newblock \emph{Language Resources and Evaluation}, 52:603 -- 623.

\bibitem[{Guo et~al.(2022)Guo, Deng, Lattas, and Zafeiriou}]{guo2022sample}
Jia Guo, Jiankang Deng, Alexandros Lattas, and Stefanos Zafeiriou. 2022.
\newblock \href {https://openreview.net/forum?id=RhB1AdoFfGE} {Sample and
  computation redistribution for efficient face detection}.
\newblock In \emph{International Conference on Learning Representations}.

\bibitem[{Hudson and Manning(2019)}]{hudson2019gqa}
Drew~A Hudson and Christopher~D Manning. 2019.
\newblock {GQA}: A new dataset for real-world visual reasoning and
  compositional question answering.
\newblock \emph{Conference on Computer Vision and Pattern Recognition (CVPR)}.

\bibitem[{Johnson et~al.(2017)Johnson, Hariharan, Van Der~Maaten, Fei-Fei,
  Lawrence~Zitnick, and Girshick}]{johnson2017clevr}
Justin Johnson, Bharath Hariharan, Laurens Van Der~Maaten, Li~Fei-Fei,
  C~Lawrence~Zitnick, and Ross Girshick. 2017.
\newblock {CLEVR}: A diagnostic dataset for compositional language and
  elementary visual reasoning.
\newblock In \emph{Proceedings of the IEEE conference on computer vision and
  pattern recognition}, pages 2901--2910.

\bibitem[{Kulkarni et~al.(2020)Kulkarni, Jain, Hosseini, Baldridge, Ie, and
  Zhang}]{Kulkarni2020SpatialLR}
Sayali Kulkarni, Shailee Jain, Mohammad~Javad Hosseini, Jason Baldridge, Eugene
  Ie, and Li~Zhang. 2020.
\newblock Spatial language representation with multi-level geocoding.
\newblock \emph{ArXiv}, abs/2008.09236.

\bibitem[{Lample et~al.(2016)Lample, Ballesteros, Subramanian, Kawakami, and
  Dyer}]{lample-etal-2016-neural}
Guillaume Lample, Miguel Ballesteros, Sandeep Subramanian, Kazuya Kawakami, and
  Chris Dyer. 2016.
\newblock \href {https://doi.org/10.18653/v1/N16-1030} {Neural architectures
  for named entity recognition}.
\newblock In \emph{Proceedings of the 2016 Conference of the North {A}merican
  Chapter of the Association for Computational Linguistics: Human Language
  Technologies}, pages 260--270, San Diego, California. Association for
  Computational Linguistics.

\bibitem[{Liu et~al.(2021)Liu, Bugliarello, Ponti, Reddy, Collier, and
  Elliott}]{liu-etal-2021-visually}
Fangyu Liu, Emanuele Bugliarello, Edoardo~Maria Ponti, Siva Reddy, Nigel
  Collier, and Desmond Elliott. 2021.
\newblock \href {https://doi.org/10.18653/v1/2021.emnlp-main.818} {Visually
  grounded reasoning across languages and cultures}.
\newblock In \emph{Proceedings of the 2021 Conference on Empirical Methods in
  Natural Language Processing}, pages 10467--10485, Online and Punta Cana,
  Dominican Republic. Association for Computational Linguistics.

\bibitem[{Lu et~al.(2020)Lu, Goswami, Rohrbach, Parikh, and Lee}]{Lu_2020_CVPR}
Jiasen Lu, Vedanuj Goswami, Marcus Rohrbach, Devi Parikh, and Stefan Lee. 2020.
\newblock 12-in-1: Multi-task vision and language representation learning.
\newblock In \emph{The IEEE/CVF Conference on Computer Vision and Pattern
  Recognition (CVPR)}.

\bibitem[{Ning et~al.(2018)Ning, Zhou, Feng, Peng, and
  Roth}]{ning-etal-2018-cogcomptime}
Qiang Ning, Ben Zhou, Zhili Feng, Haoruo Peng, and Dan Roth. 2018.
\newblock \href {https://doi.org/10.18653/v1/D18-2013} {{C}og{C}omp{T}ime: A
  tool for understanding time in natural language}.
\newblock In \emph{Proceedings of the 2018 Conference on Empirical Methods in
  Natural Language Processing: System Demonstrations}, pages 72--77, Brussels,
  Belgium. Association for Computational Linguistics.

\bibitem[{Peters et~al.(2017)Peters, Ammar, Bhagavatula, and
  Power}]{peters-etal-2017-semi}
Matthew~E. Peters, Waleed Ammar, Chandra Bhagavatula, and Russell Power. 2017.
\newblock \href {https://doi.org/10.18653/v1/P17-1161} {Semi-supervised
  sequence tagging with bidirectional language models}.
\newblock In \emph{Proceedings of the 55th Annual Meeting of the Association
  for Computational Linguistics (Volume 1: Long Papers)}, pages 1756--1765,
  Vancouver, Canada. Association for Computational Linguistics.

\bibitem[{Radford et~al.(2021)Radford, Kim, Hallacy, Ramesh, Goh, Agarwal,
  Sastry, Askell, Mishkin, Clark et~al.}]{radford2021learning}
Alec Radford, Jong~Wook Kim, Chris Hallacy, Aditya Ramesh, Gabriel Goh,
  Sandhini Agarwal, Girish Sastry, Amanda Askell, Pamela Mishkin, Jack Clark,
  et~al. 2021.
\newblock Learning transferable visual models from natural language
  supervision.
\newblock In \emph{International Conference on Machine Learning}, pages
  8748--8763. PMLR.

\bibitem[{Shen et~al.(2021)Shen, Qiu, Meng, Shang, Ren, and
  Han}]{shen-etal-2021-taxoclass}
Jiaming Shen, Wenda Qiu, Yu~Meng, Jingbo Shang, Xiang Ren, and Jiawei Han.
  2021.
\newblock \href {https://doi.org/10.18653/v1/2021.naacl-main.335}
  {{T}axo{C}lass: Hierarchical multi-label text classification using only class
  names}.
\newblock In \emph{Proceedings of the 2021 Conference of the North American
  Chapter of the Association for Computational Linguistics: Human Language
  Technologies}, pages 4239--4249, Online. Association for Computational
  Linguistics.

\bibitem[{Srinivasan et~al.(2021)Srinivasan, Raman, Chen, Bendersky, and
  Najork}]{10.1145/3404835.3463257}
Krishna Srinivasan, Karthik Raman, Jiecao Chen, Michael Bendersky, and Marc
  Najork. 2021.
\newblock \href {https://doi.org/10.1145/3404835.3463257} {\emph{{WIT}:
  Wikipedia-Based Image Text Dataset for Multimodal Multilingual Machine
  Learning}}, page 2443–2449. Association for Computing Machinery, New York,
  NY, USA.

\bibitem[{Suhr et~al.(2017)Suhr, Lewis, Yeh, and Artzi}]{suhr-etal-2017-corpus}
Alane Suhr, Mike Lewis, James Yeh, and Yoav Artzi. 2017.
\newblock \href {https://doi.org/10.18653/v1/P17-2034} {A corpus of natural
  language for visual reasoning}.
\newblock In \emph{Proceedings of the 55th Annual Meeting of the Association
  for Computational Linguistics (Volume 2: Short Papers)}, pages 217--223,
  Vancouver, Canada. Association for Computational Linguistics.

\bibitem[{Suhr et~al.(2019)Suhr, Zhou, Zhang, Zhang, Bai, and
  Artzi}]{suhr-etal-2019-corpus}
Alane Suhr, Stephanie Zhou, Ally Zhang, Iris Zhang, Huajun Bai, and Yoav Artzi.
  2019.
\newblock \href {https://doi.org/10.18653/v1/P19-1644} {A corpus for reasoning
  about natural language grounded in photographs}.
\newblock In \emph{Proceedings of the 57th Annual Meeting of the Association
  for Computational Linguistics}, pages 6418--6428, Florence, Italy.
  Association for Computational Linguistics.

\bibitem[{UzZaman et~al.(2013)UzZaman, Llorens, Derczynski, Allen, Verhagen,
  and Pustejovsky}]{UzZaman2013SemEval2013T1}
Naushad UzZaman, Hector Llorens, Leon Derczynski, James~F. Allen, Marc
  Verhagen, and James Pustejovsky. 2013.
\newblock {SemEval-2013 Task 1: TempEval-3: Evaluating Time Expressions,
  Events, and Temporal Relations}.
\newblock In \emph{*SEMEVAL}.

\bibitem[{Zhou et~al.(2020)Zhou, Ning, Khashabi, and
  Roth}]{zhou-etal-2020-temporal}
Ben Zhou, Qiang Ning, Daniel Khashabi, and Dan Roth. 2020.
\newblock \href {https://doi.org/10.18653/v1/2020.acl-main.678} {Temporal
  common sense acquisition with minimal supervision}.
\newblock In \emph{Proceedings of the 58th Annual Meeting of the Association
  for Computational Linguistics}, pages 7579--7589, Online. Association for
  Computational Linguistics.

\bibitem[{Zhou et~al.(2021)Zhou, Richardson, Ning, Khot, Sabharwal, and
  Roth}]{zhou-etal-2021-temporal}
Ben Zhou, Kyle Richardson, Qiang Ning, Tushar Khot, Ashish Sabharwal, and Dan
  Roth. 2021.
\newblock \href {https://doi.org/10.18653/v1/2021.naacl-main.107} {Temporal
  reasoning on implicit events from distant supervision}.
\newblock In \emph{Proceedings of the 2021 Conference of the North American
  Chapter of the Association for Computational Linguistics: Human Language
  Technologies}, pages 1361--1371, Online. Association for Computational
  Linguistics.

\end{thebibliography}
